\pdfoutput=1

\documentclass[screen,sigconf,authorversion]{acmart}
\usepackage{multicol}
\usepackage{tablefootnote}
\usepackage{dirtytalk}
\usepackage{breakurl}
\AtBeginDocument{%
  \providecommand\BibTeX{{%
    \normalfont B\kern-0.5em{\scshape i\kern-0.25em b}\kern-0.8em\TeX}}}

\copyrightyear{2024}
\acmYear{2024}
\setcopyright{acmlicensed}
\acmConference[ITiCSE 2024]{Proceedings of the 2024 Innovation and Technology in Computer Science Education V. 1}{July 8--10, 2024}{Milan, Italy}
\acmBooktitle{Proceedings of the 2024 Innovation and Technology in Computer Science Education V. 1 (ITiCSE 2024), July 8--10, 2024, Milan, Italy}
\acmDOI{10.1145/3649217.3653594  }
\acmISBN{979-8-4007-0600-4/   24/07}

\begin{document}

%\title{Exploring Auto-Correction of Programming Exercises With GPT-4}
\title{Feedback-Generation for Programming Exercises With GPT-4}


\author{Imen Azaiz}
\email{imen.azaiz@ifi.lmu.de}
\orcid{0009-0005-6458-4169}
\affiliation{%
	\institution{LMU Munich}
	\department{}
	\streetaddress{}
	\city{Munich}
	\country{Germany}
}

\author{Natalie Kiesler}
\email{natalie.kiesler@th-nuernberg.de}
\orcid{0000-0002-6843-2729}
\affiliation{%
 \institution{Nuremberg Tech}
 \streetaddress{}
 \city{Nuremberg}
 \country{Germany}}

\author{Sven Strickroth}
\email{sven.strickroth@ifi.lmu.de}
\orcid{0000-0002-9647-300X}
\affiliation{%
	\institution{LMU Munich}
	\department{}
	\streetaddress{}
	\city{Munich}
	\country{Germany}
}

%\renewcommand{\shortauthors}{Anonymous, et al.}

\begin{abstract}
Ever since Large Language Models (LLMs) and related applications have become broadly available, several studies investigated their potential for assisting educators and supporting students in higher education. LLMs such as Codex, GPT-3.5, and GPT 4 have shown promising results in the context of large programming courses, where students can benefit from feedback and hints if provided timely and at scale. This paper explores the quality of GPT-4 Turbo's generated output for prompts containing both the programming task specification and a student's submission as input. Two assignments from an introductory programming course were selected, and GPT-4 was asked to generate feedback for 55 randomly chosen, authentic student programming submissions. The output was qualitatively analyzed regarding correctness, personalization, fault localization, and other features identified in the material. Compared to prior work and analyses of GPT-3.5, GPT-4 Turbo shows notable improvements. For example, the output is more structured and consistent. GPT-4 Turbo can also accurately identify invalid casing in student programs' output. In some cases, the feedback also includes the output of the student program. At the same time, inconsistent feedback was noted such as stating that the submission is correct but an error needs to be fixed. The present work increases our understanding of LLMs' potential, limitations, and how to integrate them into e-assessment systems, pedagogical scenarios, and instructing students who are using applications based on GPT-4.
\end{abstract}






\maketitle

\section{Introduction} 

Large Language Models (LLMs) not only took the world by storm, they also have a potentially great impact on programming education comprising both opportunities and challenges for learners and educators~\cite{prather2023wgfullreport}. With the release of OpenAI's new model GPT-4 Turbo in November 2023, an updated knowledge cutoff until April 2023, and an increased context window were made available along with the extension to become a multimodal model~\cite{openaigpt4}. The new model yet again raises the question of to what extent generative AI tools can be used to create truly individual, reliable feedback that is adequate for novice learners of programming.

Individual feedback may help counteract the challenges well-known in the context of introductory programming classes and improve student performance, if addressing students' (informational) needs~\cite{narciss2008feedback,shute,hao2022towards}. 
Novice learners of programming usually face several challenges in the introductory phase of their studies.
Programming may be completely new to them due to their educational biography, and it is considered a cognitively complex tasks~\cite{kiesler2020towardsiticse,Kiesler2024}, involving cognitive challenges (e.\,g., problem understanding, developing algorithms, debugging, understanding error messages~\cite{duboulay1986some,Luxton-Reilly2018,Ebert2016,spohrer1986novice}). Moreover, expectations from educators and institutions towards students seem to be too high and unrealistic~\cite{luxton-reilly2016,Luxton-Reilly2018,whalley2007many}. 

At the same time, educators struggle with high student numbers, limited resources to provide feedback and hints, lack of tutors, and an overall heterogeneity of their students~\cite{petersen2016revisiting,SB22}. Considering the scenarios of (ungraded) formative assessments to support students' learning process, receiving feedback is key for them to improve their work during the semester. It is therefore not surprising that many (formative) e-assessment systems have been developed to support both educators and students~\cite{Keuning2018,jeuring2022towards,SS22}. Yet, many of the current systems only focus on functioning code and automatic tests instead of individualized feedback. 

In this paper, \textbf{the goal} is to explore the capabilities of GPT-4 Turbo to generate formative feedback for programming exercises. The research question (RQ) is \textit{How can we characterize the feedback provided by GPT-4 Turbo if provided with a task description and a student solution as input?} By answering this RQ, we contribute to the body of research improving the computing education research community's understanding of LLMs and their potential benefits for novice programmers. We also discuss the use case of applying the GPT-4 API as part of a university's e-assessment system. 

Therefore, this work has implications for educators and e-assess\-ment system developers considering the integration of the GPT-4 API into their courses or systems. As ChatGPT is based on GPT-4, this research also has implications for students seeking help for a certain issue, and educators instructing students on the conscious and critical use of GPT-4's feedback. 

\section{Related Research}

In the past decades, numerous e-assessment systems, intelligent tutoring systems (ITS), and learning environments have been developed to provide automatic feedback and hints to students~\cite{SS22,Keuning2018,jeuring2022towards}. These systems can provide timely feedback at scale without the need for an educator's intervention. For most systems, test cases are automatically executed upon a student's submission to generate feedback~\cite{Keuning2018}. A precondition for this to work is the development of such tests (plus a domain model in case of an ITS), causing tremendous effort for educators. Some systems also employ professional code analysis tools such as PMS, CheckStyle, or SonarCube to provide feedback on style~(e.\,g., \cite{Liu2020}). A common problem is, however, that the provided feedback is not always useful for students, as the descriptions lack details on how to proceed~\cite{Becker2019,Keuning2021}. 

In the context of ITS, using AI has a long-standing tradition \cite{TUC-HCIS/LSGP13}. More recently, LLMs have become widely available and are being explored for application as feedback generators for novice learners of programming. 
Several papers including a recent ITiCSE working group report~\cite{prather2023wgfullreport,denny2023computing,becker2023generative} discuss implications of recent AI advances for computing education, particularly for introductory programming courses. Early papers used OpenAI's Codex model and demonstrated its ability to solve CS1 and CS2 programming tasks at a level similar to students. Occasionally, it faced difficulties with output formatting, odd edge cases, ambiguous requirements, and wordy tasks~\cite{Finnie2022,FinnieAnsley2023MyAW}.

LLMs have also shown be be able to create effective code explanations as well as enhance programming error messages~\cite{Leinonen2023,Sarsa2022,macneil2022experiences}. \citeauthor{macneil2022experiences} conclude that the majority of students perceived the automatically generated line-by-line code explanations by an LLM as helpful when evaluated as part of an e-book~\cite{macneil2022experiences}. \citeauthor{leinonen2023comparing} conclude that code explanations generated by GPT-3 are rated better on average w.r.t. understanding and accuracy than explanations created by students. Moreover, students were not averse to feedback generated by LLMs, and they prefer line-by-line explanations~\cite{leinonen2023comparing}.

%Absatz bearbeiten, sodass der nicht auf nächste Spalte umbricht
Considering the context of LLM-generated feedback (GPT-4) to programming problems and how to elicit it, the accuracy of the feedback seems to improve when the model receives the task instruction as input~\cite{Bengtsson_Kaliff_2023}. At the same time, such input seems to cause a decreasing LLM performance in identifying errors, indicating the need for more research~\cite{Bengtsson_Kaliff_2023}. A study on the generation of next-step hints by GPT-3.5~\cite{roest2023nextstep} recommends the use of the task description and keywords as input. Regardless of the input, LLM-generated feedback messages may contain misleading information~\cite{kiesler2023exploring} and lack sufficient detail when students approach the end of the assignment~\cite{roest2023nextstep}. \citeauthor{LMU-TEL/ADS2023} note further difficulties of GPT-3.5 with output formatting, hallucinating errors, and recognizing correct solutions, resulting in adequate feedback in only 47\,\% of the cases.

Other recent studies on GPT-4 investigate its use to localize errors in program code~\cite{wu2023large}, and to what extent it passes assessments from introductory and intermediate Python classes~\cite{savelka2023large}. However, the latest version of OpenAI's LLM (GPT-4 Turbo)~\cite{openaigpt4} has not yet been subject to qualitative research regarding its feedback capabilities. Moreover, the breadth and depth to which we explore the feedback structure and quality in the context of introductory programming is a novel contribution to the computing education research community.

\section{Methodology}

The goal of this work is to explore the formative feedback generation capabilities of GPT-4 Turbo for introductory programming exercises. Our research is guided by the following RQ: \textit{How can we characterize the feedback provided by GPT-4 Turbo if provided with a task description and a student solution as input?}

To evaluate the feedback generated by GPT-4, a qualitative empirical study was conducted. Two assignments from an introductory Java course at LMU Munich, a large German university, were selected along with authentic student data. The authors obtained and reused available student data from related work~\cite{LMU-TEL/ADS2023}. The dataset contains all submissions from 695 computer science students (majors and minors), who took a first-year introductory programming class in the winter term 2021/22. The course was accompanied by weekly homework assignments and peer reviews. 
Participation was voluntary. The e-assessment system GATE~\cite{SOP11} was used to collect the submissions, to provide instant feedback for some tasks, and to facilitate the peer review process~\cite{Strickroth2023}.
This research utilizes submissions from students who explicitly consented to their use (695 out of about 900 students). 
Consent was fully voluntary, with no negative consequences or disadvantages. 

The first task we selected from the dataset required students to: \say{Write a Java application named \textit{SimpleWhileLoop} that uses a \textit{WHILE} loop to count and prints all odd numbers from 1 to 10, and then prints `Boom!' (without quotation marks) afterward.} It was expected in week 2 of the class. 
The second task we selected was due in week 7. It is object-oriented and expects students to implement an interface, use an inner-class, write multiple methods, traverse linked lists, and manage references. The assignment specification is: \say{Implement the Queue interface according to the specification (in the interface) for a queue with the QueueImpl class by using a singly linked list.}
The Java interface \textit{Queue} was provided as a Java file containing the following five methods with JavaDoc specifying the semantics: \textit{void append(int value)}, \textit{boolean isEmpty()}, \textit{void remove()} (null operation for an queue list), \textit{int peek()} (\textit{EMPTY\_VALUE} of the interface should be returned on an empty queue), and \textit{int[] toArray()}.
%public interface Queue {
%    final int EMPTY_VALUE = -1;
%
%    /**
%     * Appends the value (at the end) to the queue.
%     * @param value
%     */
%    void append(int value);
%
%    /**
%     * Checks if the queue is empty.
%     * @return true if the queue is empty
%     */
%    boolean isEmpty();
%
%    /**
%     * Removes the first element from the queue.
%     * Does nothing if the queue is empty.
%     */
%    void remove();
%
%    /**
%     * Returns the value of the first element in the queue.
%     * If the queue is empty, it returns EMPTY_VALUE.
%     * @return Value of the first element
%     */
%    int peek();
%
%    /**
%     * Creates an array with the values of the queue.
%     * @return a fresh array with the values of the queue.
%     */
%    int[] toArray();
%}

About 9\,\% of all submissions for these two assignments were used in this study: 
33 submissions were pseudo-randomly sampled for the first assignment and 22 randomly for the second one. During our review of the selected solutions, we found that two submissions for the second task were very similar, i.\,e. only differing w.r.t. the extensive use of comments. We kept both due to their authenticity.

For the generation of feedback, we used the GPT-4 Turbo (gpt-4-1106-preview) model with default settings and the following prompt template, following the methodology in related work~\cite{LMU-TEL/ADS2023}:
\small
\begin{verbatim}
[ASSIGNMENT INSTRUCTIONS]
Find all kinds of errors, including logical ones, and
provide hints for their correction or improvement,
including suggestions for code style.
[CODE OF STUDENT SUBMISSION]
\end{verbatim}
\normalsize

We experimented with several variations of the prompt but could not find significant differences.
Next, the feedback was generated three times for every submission in randomized order using the very same model and configuration (zero-shot approach). The submissions of the first assignment were processed on November, 21st 2023 resulting in 99 feedback texts. The submissions of the second task were processed on January, 4th 2024, resulting in 84 outputs. 

All outputs were manually analyzed using a qualitative thematic analysis technique~\cite{Mayring2001,braun2006using}. The classification in related work \cite{LMU-TEL/ADS2023} was used as a starting point for the deductive-inductive category-building process. Inductive categories were developed based on the material to describe new feedback characteristics.  
Three computing education researchers with extensive expertise in correction and providing feedback as well as qualitative analysis were involved in the coding and development of the categories.

In addition to the thematic analysis, all submissions were manually checked using unit tests for syntactic and functional correctness. Functional correctness assumes the submission fulfills the task specification and works as expected (i.\,e., regardless of its performance).
Moreover, we evaluate GPT's accuracy, precision, and recall. 
For all categories representing feedback characteristics, we counted the frequencies related to either of the tasks.

\section{Results}
In this section, we present the results of the investigation of the feedback generated by GPT-4 Turbo. We characterized the feedback based on its form (e.\,g., content, structure, length, and overall composition), and evaluated the correctness of the feedback before examining the types of corrections provided by the model. Moreover, the results reflect code optimizations, style recommendations, inconsistencies, and redundancies. Variations of the feedback characteristics depending on the assignments are discussed related to these five main characteristics. Table~\ref{tab:codingbook} represents the codebook, whereas examples are provided in the text, where appropriate.

\begin{table}[h!]
\small
\caption{Coding book with descriptions (examples are provided in the text where appropriate)}\label{tab:codingbook}
\begin{tabular}{|p{2.2cm}|p{5.7cm}|}
\hline
\textbf{Category} & \textbf{Description} \\
\hline
\multicolumn{2}{|l|}{\textit{Feedback Content and Structure}} \\
\hline
Feedback without code (FWOC) & Feedback contains plain text without code (lacking Java programming language keywords or variable/\allowbreak{}method names). \\
\hline
Feedback text with code (FTWC) & Feedback contains text with code, snippet, variable/method name. \\
\hline
Feedback just containing code (FJCC) & Feedback contains only code. \\
\hline
Compliance with spec. (CWAS) & Corrections or suggestions align with the provided instructions and assignment specification.  \\
\hline  
\multicolumn{2}{|l|}{\textit{Code Representation}}\\
\hline
Full code (FuCo)&
Suggests a full program sample solution.\\
\hline
Code snippet (CoSn) &
Corrects small portions of the program suggesting a sequence of instructions. \\
\hline
Code snippet with instruction (CoSnI) & Generates code snippets with gaps, including instructions for students on how to fill in the remaining gaps. \\
\hline
Code with output (CWO) & Suggests improvements in the code with the corresponding output. \\
\hline
Inline code correction (ICC) & Feedback text contains student solution with inline comments (corrections and suggestions). \\
\hline
\multicolumn{2}{|l|}{\textit{Correctness and Correction Types}}\\
\hline
Only correct cor\-rec\-tion/\allowbreak{}sug\-ges\-tions (OCCS) & Feedback contains only correct improvements/\allowbreak{}sug\-gesti\-ons, meaning all contained errors were fixed. Moreover, all of the suggestions have been implemented, resulting in the display of running code. \\
\hline 
Partially correct cor\-rec\-tion/\allowbreak{}sug\-ges\-tion (PCCS) & Only some feedback components are correct, while other components introduce new issues (i.\,e., incorrect feedback or suggestions).\\
\hline
Only false correction/suggestion (OFCS) &
Feedback contains only false corrections like non-existent errors or suggestions resulting in broken code.\\
\hline
Completely correct correction (CCC) &
Feedback addresses all of the submitted code's issues, contains only correct corrections, and adheres to the task requirements. Applying the feedback results in a fully correct submission. \\
\hline
(Fault) localization (FL) & 
Bugs are identified and localized, e.\,g., by citing code snippets, or describing them. \\
\hline
(Fault) localization correct (FLC) & 
Bugs are correctly identified and localized and are present in these locations.\\
\hline
\multicolumn{2}{|l|}{\textit{Suggested Optimizations and Coding Style}}\\
\hline
Optimization (OPT) & Suggests optimizations regarding the functionality of the program.\\
\hline
Code style suggestion (CSS) & Suggests improvements regarding readability, documentation, comments within the code, variable naming, etc. \\
\hline
Language sug\-ges\-tion (LCS) & 
Feedback contains translations and language related suggestions. \\
\hline
\multicolumn{2}{|l|}{\textit{Inconsistencies and Redundancies}}\\
\hline
Inconsistency (InC) & Recommendation does not correspond to the sample solution, or contradiction within the textual feedback. \\
\hline
Redundancy (RD) & Repeats the same suggestion in the same feedback or provides a suggestion that is already implemented in the code. \\
\hline
\end{tabular}
\end{table}

\subsection{Feedback Content, Structure, and Length}
The deductive-inductive characterization of the feedback generated by GPT-4 started with the development and application of categories reflecting its content and structure (see \autoref{tab:codingbook}). We further analyzed the length of the responses.

\subsubsection{Content and Structure}
Regarding the content of the generated feedback, we found that 100\,\% of the output contained both code and text (FTWC) as shown in the first section of \autoref{tab:selected_properties}. 
Overall, we found the content of the generated output to be \say{individualized} to the input, meaning there were very few repetitive elements in all of the LLMs' responses. 
The content was almost always compliant with the assignment (in 98\,\%). In only one response to a student solution for the \textit{SimpleWhileLoop}, the odd numbers followed by the word \say{Boom} were not displayed line-by-line in any of the three iterations. Yet, GPT-4 Turbo acknowledged this issue in the textual output. 

In general, the feedback generated by GPT-4 Turbo seemed to exhibit a certain structure, usually comprising sections. 
The first part is an introductory statement or a description of the submitted code. This is coupled with an assessment of the code's quality and correctness. 
Next, an (enumerated) list of issues and respective corrections or suggestions for improvements were mostly displayed. These were accompanied by the improved version of the complete code (FuCo) or code snippet (CoSn). 
Usually, these list items were categorized under various labels such as \say{Logical Errors}, \say{Corrections and Improvements}, \say{Code Style and Clarity}, \say{Code Efficiency}, \say{Error Handling}, or \say{Variable Naming}. 
As a last part, GPT-4 Turbo generated a summary (of its corrections) along with final remarks.

The order of hints and corrections, however, often seems random, e.\,g. coding style hints are provided before errors, recommendations for encapsulation are not bundled, or a missing inner class is provided as the last point. 
Another observation is the repetition of several text fragments across the responses. Those were, for example, related to code style: \say{Always use curly braces `\{\}' for blocks under `if' statements and loops}.

\begin{table}[!htbp]
\caption{Frequencies of all codes applied to both tasks}\label{tab:selected_properties}
\centering
\small
\begin{tabular}{crrrrrrrr|}
\cline{2-9}
\multicolumn{1}{c|}{} & \multicolumn{3}{c|}{\begin{tabular}[c]{@{}c@{}}\textit{SimplieWhileLoop} \\ n =33\end{tabular}} & \multicolumn{3}{c|}{\begin{tabular}[c]{@{}c@{}}\textit{Queue}\\  n=22\end{tabular}} & \multicolumn{2}{c|}{\begin{tabular}[c]{@{}c@{}}All \\ n=165\end{tabular}} \\ \hline
\multicolumn{1}{|c|}{Char.} & \multicolumn{1}{r|}{1st} & \multicolumn{1}{r|}{2nd} & \multicolumn{1}{r|}{3rd} & \multicolumn{1}{r|}{1st} & \multicolumn{1}{r|}{2nd} & \multicolumn{1}{r|}{3rd} & \multicolumn{1}{r|}{Sum} & \multicolumn{1}{r|}{\%} \\ \hline
\multicolumn{9}{|l|}{\textit{Feedback  Content and Structure}} \\ \hline
\multicolumn{1}{|c|}{FWOC} & \multicolumn{1}{r|}{0} & \multicolumn{1}{r|}{0} & \multicolumn{1}{r|}{0} & \multicolumn{1}{r|}{0} & \multicolumn{1}{r|}{0} & \multicolumn{1}{r|}{0} & \multicolumn{1}{r|}{0} & 0 \\ \hline
\multicolumn{1}{|c|}{FTWC} & \multicolumn{1}{r|}{33} & \multicolumn{1}{r|}{33} & \multicolumn{1}{r|}{33} & \multicolumn{1}{r|}{22} & \multicolumn{1}{r|}{22} & \multicolumn{1}{r|}{22} & \multicolumn{1}{r|}{165} &  100 \\ \hline
\multicolumn{1}{|c|}{FJCC} & \multicolumn{1}{r|}{0} & \multicolumn{1}{r|}{0} & \multicolumn{1}{r|}{0} & \multicolumn{1}{r|}{0} & \multicolumn{1}{r|}{0} & \multicolumn{1}{r|}{0} & \multicolumn{1}{c|}{0} & 0 \\ \hline
\multicolumn{1}{|c|}{CWAS} & \multicolumn{1}{r|}{32} & \multicolumn{1}{r|}{32} & \multicolumn{1}{r|}{32} & \multicolumn{1}{r|}{22} & \multicolumn{1}{r|}{22} & \multicolumn{1}{r|}{22} & \multicolumn{1}{r|}{162} &  98 \\ \hline
\multicolumn{9}{|l|}{\textit{Code Representation}} \\ \hline
\multicolumn{1}{|c|}{FuCo} & \multicolumn{1}{r|}{33} & \multicolumn{1}{r|}{33} & \multicolumn{1}{r|}{33} & \multicolumn{1}{r|}{12} & \multicolumn{1}{r|}{10} & \multicolumn{1}{r|}{11} & \multicolumn{1}{r|}{132} &  80 \\ \hline
\multicolumn{1}{|c|}{CoSn} & \multicolumn{1}{r|}{1} & \multicolumn{1}{r|}{3} & \multicolumn{1}{r|}{2} & \multicolumn{1}{r|}{10} & \multicolumn{1}{r|}{14} & \multicolumn{1}{r|}{14} & \multicolumn{1}{r|}{44} & 27  \\ \hline
\multicolumn{1}{|c|}{CoSnI} & \multicolumn{1}{r|}{0} & \multicolumn{1}{r|}{0} & \multicolumn{1}{r|}{0} & \multicolumn{1}{r|}{1} & \multicolumn{1}{r|}{0} & \multicolumn{1}{r|}{1} & \multicolumn{1}{r|}{2} & 1 \\ \hline
\multicolumn{1}{|r|}{CWO} & \multicolumn{1}{c|}{5} & \multicolumn{1}{r|}{1} & \multicolumn{1}{r|}{5} & \multicolumn{1}{r|}{0} & \multicolumn{1}{r|}{0} & \multicolumn{1}{r|}{0} & \multicolumn{1}{r|}{11} &  7\\ \hline
\multicolumn{1}{|c|}{ICC} & \multicolumn{1}{r|}{7} & \multicolumn{1}{r|}{3} & \multicolumn{1}{r|}{3} & \multicolumn{1}{c|}{0} & \multicolumn{1}{r|}{0} & \multicolumn{1}{r|}{0} & \multicolumn{1}{r|}{13} & 8 \\ \hline
\multicolumn{9}{|l|}{\textit{Correctness and Correction Types}} \\ \hline
\multicolumn{1}{|c|}{OCCS} & \multicolumn{1}{r|}{21} & \multicolumn{1}{r|}{23} & \multicolumn{1}{r|}{19} & \multicolumn{1}{r|}{13} & \multicolumn{1}{r|}{10} & \multicolumn{1}{r|}{13} & \multicolumn{1}{r|}{99} & 60 \\ \hline
\multicolumn{1}{|c|}{PCCS} & \multicolumn{1}{r|}{12} & \multicolumn{1}{r|}{10} & \multicolumn{1}{r|}{14} & \multicolumn{1}{r|}{9} & \multicolumn{1}{r|}{12} & \multicolumn{1}{r|}{9} & \multicolumn{1}{r|}{66} & 40 \\ \hline
\multicolumn{1}{|c|}{OFCS} & \multicolumn{1}{r|}{0} & \multicolumn{1}{r|}{0} & \multicolumn{1}{r|}{0} & \multicolumn{1}{c|}{0} & \multicolumn{1}{c|}{0} & \multicolumn{1}{r|}{0} & \multicolumn{1}{r|}{0} & 0 \\ \hline
\multicolumn{1}{|c|}{CCC} & \multicolumn{1}{r|}{21} & \multicolumn{1}{r|}{19} & \multicolumn{1}{r|}{18} & \multicolumn{1}{r|}{11} & \multicolumn{1}{r|}{8} & \multicolumn{1}{r|}{9} & \multicolumn{1}{r|}{86} & 52 \\ \hline
\multicolumn{1}{|c|}{FL} & \multicolumn{1}{r|}{18} & \multicolumn{1}{r|}{19} & \multicolumn{1}{r|}{20} & \multicolumn{1}{r|}{21} & \multicolumn{1}{r|}{21} & \multicolumn{1}{r|}{22} & \multicolumn{1}{r|}{121} & 73 \\ \hline
\multicolumn{1}{|c|}{FLC} & \multicolumn{1}{r|}{16} & \multicolumn{1}{r|}{13} & \multicolumn{1}{r|}{17} & \multicolumn{1}{r|}{19} & \multicolumn{1}{r|}{21} & \multicolumn{1}{r|}{20} & \multicolumn{1}{r|}{106} & 64 \\ \hline
\multicolumn{9}{|l|}{\textit{Suggested Optimizations and Coding Style}} \\ \hline
\multicolumn{1}{|c|}{OPT} & \multicolumn{1}{r|}{19} & \multicolumn{1}{r|}{10} & \multicolumn{1}{r|}{5} & \multicolumn{1}{r|}{19} & \multicolumn{1}{r|}{21} & \multicolumn{1}{r|}{22} & \multicolumn{1}{r|}{96} & 58 \\ \hline
\multicolumn{1}{|c|}{CSS} & \multicolumn{1}{r|}{33} & \multicolumn{1}{r|}{33} & \multicolumn{1}{r|}{33} & \multicolumn{1}{r|}{19} & \multicolumn{1}{r|}{21} & \multicolumn{1}{r|}{20} & \multicolumn{1}{r|}{159} & 96 \\ \hline
\multicolumn{1}{|c|}{LCS} & \multicolumn{1}{r|}{6} & \multicolumn{1}{r|}{7} & \multicolumn{1}{r|}{6} & \multicolumn{1}{r|}{5} & \multicolumn{1}{r|}{4} & \multicolumn{1}{r|}{2} & \multicolumn{1}{r|}{30} & 18 \\ \hline
\multicolumn{9}{|l|}{\textit{Inconsistencies and Redundancies}} \\ \hline
\multicolumn{1}{|c|}{InC} & \multicolumn{1}{r|}{6} & \multicolumn{1}{r|}{6} & \multicolumn{1}{r|}{7} & \multicolumn{1}{r|}{6} & \multicolumn{1}{r|}{4} & \multicolumn{1}{r|}{8} & \multicolumn{1}{r|}{37} & 22 \\ \hline
\multicolumn{1}{|c|}{RD} & \multicolumn{1}{r|}{3} & \multicolumn{1}{r|}{1} & \multicolumn{1}{r|}{2} & \multicolumn{1}{r|}{6} & \multicolumn{1}{r|}{5} & \multicolumn{1}{r|}{0} & \multicolumn{1}{r|}{17} & 10 \\ \hline
\end{tabular}
\end{table}

\subsubsection{Length}
In addition to the construction of inductive categories, we analyzed the length of the responses, as they were extensive. \autoref{tab:wordcountall} shows the word counts for both assignments and across all three runs. The number of words was determined by tokenizing the feedback string using the white space (\verb|"\s+"|) and counting the resulting tokens. The overall median feedback length is $m=360$ words ($\bar{x}=381$). It seems to be quite consistent across the three runs. The generated feedback for the \textit{SimpleWhileLoop} assignment ($m=312$) is shorter than for the \textit{Queue} ($m=470$). This difference is statistically significant (Mann-Whitney UTest, $U=168$, $p<.001$, two-sided).

\begin{table}[!htb]
\centering
\smaller
\caption{Length of the generated feedback by word counts (OA: over all iterations for each assignment)}
\label{tab:wordcountall}
\begin{tabular}{c|rrrr|cccr|r}
\cline{2-9}
 & \multicolumn{4}{c|}{SimpleWhileLoop} & \multicolumn{4}{c|}{Queue} &  \\ \hline
\multicolumn{1}{|c|}{Words} & \multicolumn{1}{c|}{1st} & \multicolumn{1}{c|}{2nd} & \multicolumn{1}{c|}{3rd} & OA & \multicolumn{1}{c|}{1st} & \multicolumn{1}{c|}{2nd} & \multicolumn{1}{c|}{3rd} & OA & \multicolumn{1}{c|}{All} \\ \hline
\multicolumn{1}{|c|}{Mean} & \multicolumn{1}{c|}{309} & \multicolumn{1}{c|}{310} & \multicolumn{1}{c|}{324} & 315 & \multicolumn{1}{c|}{466} & \multicolumn{1}{c|}{504} & \multicolumn{1}{c|}{477} & 482 & \multicolumn{1}{c|}{382} \\ \hline
\multicolumn{1}{|c|}{Median} & \multicolumn{1}{c|}{309} & \multicolumn{1}{c|}{306} & \multicolumn{1}{c|}{325} & 312 & \multicolumn{1}{c|}{456.5} & \multicolumn{1}{c|}{508} & \multicolumn{1}{c|}{477.5} & 470 & \multicolumn{1}{c|}{360} \\ \hline
\multicolumn{1}{|c|}{Min} & \multicolumn{1}{c|}{219} & \multicolumn{1}{c|}{223} & \multicolumn{1}{c|}{168} & 168 & \multicolumn{1}{c|}{357} & \multicolumn{1}{c|}{409} & \multicolumn{1}{c|}{339} & 339 & \multicolumn{1}{c|}{168} \\ \hline
\multicolumn{1}{|c|}{Max} & \multicolumn{1}{c|}{407} & \multicolumn{1}{c|}{423} & \multicolumn{1}{c|}{483} & 483 & \multicolumn{1}{c|}{579} & \multicolumn{1}{c|}{610} & \multicolumn{1}{c|}{619} & 619 & \multicolumn{1}{c|}{619} \\ \hline
\end{tabular}
\end{table}

\subsection{Code Representation} 
The representation of code was another theme we identified in the responses in varying forms. 
The second section of \autoref{tab:selected_properties} highlights the variation of the feedback representing, for example, the full code (FuCo) and code snippets (CoSn) as suggestions. Specifically, every feedback for the \textit{SimpleWhileLoop} included full code. Only three occurrences of code snippets were identified. 
In contrast to that, about half of the feedback generated for the \textit{Queue} task contained full code. The other half only contained code snippets.

The other three characteristics related to the code's representation also varied depending on the assignment. Code snippets with instructions (CoSnI) were provided exclusively in the \textit{Queue} assignment's feedback. This usually took the form of a code snippet with guidance for the student on how to continue. 
Inline code corrections (ICC), and code paired with its corresponding output (CWO) were only generated in response to the \textit{SimpleWhileLoop}, appearing in 13 and 11 outputs respectively.

\subsection{Feedback Correctness and Correction Types}
\label{sec:fbcorr}

Before characterizing the LLMs' output correctness and their correction types, we briefly indicate the quality of the students' code and their errors. The majority (57\,\%) of the student solutions for the \textit{SimpleWhileLoop} is fully correct (with 90\,\% having syntactic correct), whereas 5\,\% of the \text{Queue} submissions are fully correct (with 59\,\% containing syntactic correct).

Evaluating GPT's classification performance in finding errors was used as a starting point, before constructing inductive categories to describe the output's correctness and types of corrections. When we started to explore the correctness of GPT's feedback to students' submissions, it was often not explicit whether GPT had classified a submission as correct or incorrect. In the absence of explicit judgments, we used terms such as \say{error} and \say{correction} to develop categories reflecting GPT's corrections. It is important to note the difference between corrections and suggestions for improvement. We thus did not categorize suggestions regarding code style and optimization as errors.

\autoref{tab:metrics} shows the results of the analysis w.r.t. GPT's output, and its correctness. Overall, the accuracy (i.\,e., ratio of correct results to all results) ranges between .75 and .81 for the \textit{SimpleWhileLoop} and between .9 and .95 for the \textit{Queue}. The precision (i.\,e., ratio of correct positive results to all positive results) is optimal for the \textit{Queue} and between .78 and .91 for the \textit{SimpleWhileLoop}. However, the recall (i.\,e., ratio of correct positive results to all actual positive results) is better for the \textit{SimpleWhileLoop} (.68). For the \textit{Queue} task, it ranges between .33 and .66.

\begin{table}[!htb]
\centering
\smaller
\caption{Comparison of evaluation metrics of GPT-4 Turbo’s classification performance
across the three runs for the two assignments}
\label{tab:metrics}
\begin{tabular}{c|rrrr|rrrr|r}
\cline{2-9}
 & \multicolumn{4}{c|}{SimpleWhileLoop} & \multicolumn{4}{c|}{Queue} & \multicolumn{1}{l}{} \\ \hline
\multicolumn{1}{|c|}{Metric} & \multicolumn{1}{c|}{1st} & \multicolumn{1}{c|}{2nd} & \multicolumn{1}{c|}{3rd} & \multicolumn{1}{c|}{OA} & \multicolumn{1}{c|}{1st} & \multicolumn{1}{c|}{2nd} & \multicolumn{1}{c|}{3rd} & \multicolumn{1}{c|}{OA} & \multicolumn{1}{c|}{All} \\ \hline
\multicolumn{1}{|c|}{Accuracy} & \multicolumn{1}{r|}{.75} & \multicolumn{1}{r|}{.81} & \multicolumn{1}{r|}{.75} & .77 & \multicolumn{1}{r|}{.95} & \multicolumn{1}{r|}{.90} & \multicolumn{1}{r|}{.90} & .91 & \multicolumn{1}{r|}{.84} \\ \hline
\multicolumn{1}{|c|}{Precision} & \multicolumn{1}{r|}{.78} & \multicolumn{1}{r|}{.91} & \multicolumn{1}{r|}{.78} & .82 & \multicolumn{1}{r|}{1.00} & \multicolumn{1}{r|}{1.00} & \multicolumn{1}{r|}{1.00} & 1.00 & \multicolumn{1}{r|}{.91} \\ \hline
\multicolumn{1}{|c|}{Recall} & \multicolumn{1}{r|}{.68} & \multicolumn{1}{r|}{.68} & \multicolumn{1}{r|}{.68} & .68 & \multicolumn{1}{r|}{.66} & \multicolumn{1}{r|}{.33} & \multicolumn{1}{r|}{.33} & .44 & \multicolumn{1}{r|}{.56} \\ \hline
\end{tabular}
\end{table}

Most of the generated feedback texts contained correct corrections (OCCS, 60\,\%). 
No feedback contained false corrections only. However, in 40\,\% of the feedback the corrections were only partially correct (PCCS), meaning some of the feedback was incorrect (see \autoref{tab:codingbook}). To localize errors, GPT-4 cites or highlights code, which we found in 73\,\% of the output (88\,\% of these errors were correct).

In 52\,\% of the outputs, the corrections were complete (CCC), meaning the feedback contained only correct corrections and suggestions, met the task requirements, and all issues of the student solution were addressed as part of the correction. In general, applying the corrections -- even those with an incorrect explanation or reason -- always resulted in a correct solution, except for two cases. In these two cases, the package was not corrected, but GPT-4 expressed that the package may needs to be changed.

Student errors in the \textit{SimpleWhileLoop} were related to the capitalization of the word \say{Boom!} and the loop. All of these capitalization errors were identified and corrected by the LLM. In a few cases, however, GPT-4 identified a \say{logical error} when the loop variable was not initialized with 1, loop-conditions were too complex, or unnecessary if-conditions were included in the student's code.

A common student error in the \textit{Queue} task was that the inner class for the \textit{Node} was missing. This way always detected and corrected by GPT-4. Two student submissions had re-used an \textit{ArrayList} or \textit{LinkedList} (not singly-linked), which was also detected by GPT-4.
GPT-4 also noted potential memory leaks (e.\,g., \textit{tail} was not reset) in functional correct submissions and highlighted these as an error. 

There are several cases in the \textit{Queue} output where an error explanation is not correct, but the correction is. For example, GPT-4 correctly spotted an error in the \textit{toArray} method. However, it reported an \textit{ArrayIndexOutOfBoundsException}, whereas the actual error is a \textit{NullPointerException}.
In another example, a closing comment was not recognized as missing. GPT-4 also had problems in two runs with an inner class, which was named the same as the interface (Java allows such a scenario). 
Other issues occurred related to default initialized member variables. A missing explicit initialization in the constructor was often reported as an error. 
We also found problematic components of corrections (PCCS) such as \say{return an empty array not array with 0 length} or \say{The `remove' method [\dots] doesn't dispose of [sic] the removed object}. This is not correct for Java and may be misleading for students. 

\subsection{Suggested Optimizations and Coding Style}
Most of the generated feedback texts (56\,\%) contained suggestions for optimizations (OPT, see \autoref{tab:codingbook}). These can be characterized as either prescriptive \textit{should do} and discretionary \textit{could do}.
Most optimizations relate to performance, such as incrementing a variable by two instead of two increments by one, or introducing a \textit{tail} reference, and adding the field \textit{size}.
Further optimization \mbox{suggestions} (OPT) relate to discarding a (redundant) \textit{size} field (with problematic time-complexity arguments), simplifying if-conditions, encapsulation, adding @Override, and avoiding to print errors to the console. 
None of the suggested prescriptive optimizations resulted in an error or violation of the task specification. 
However, multiple aspects of the feedback may be too complex for students. This is particularly true for the \textit{Queue} task, and concerns the handling of integer overflow, or taking care of concurrent access. Only one of the four provided concurrency fixes was correct. 
Moreover, we identified several inaccurate aspects, such as GPT suggesting to put the provided code for a private and/or static inner class into a separate file.
In addition, some of the discretionary suggestions go beyond the task specification, as they recommend improving exception handling or using Generics.

Almost all of the generated feedback (96\,\%) contains code style suggestions (CSS). Most suggestions relate to the naming of variable/\allowbreak{}method/\allowbreak{}class (e.\,g., fixing typos, suggesting better names or CamelCase). Further suggestions encompass consistently using braces, proper indentation, using the interface constant instead of a literal, making variables final, avoiding redundant else blocks, deleting redundant comments, and only using \textit{this} if necessary.

Regarding the language (LCS), GPT-4 provided feedback when comments or variable names were in German. The LLM suggested translating these to English, which we found in 18\,\% of the outputs.

\subsection{Inconsistencies and Redundancies}

%inconsistencies
Feedback inconsistencies (InC) are defined as recommendations that do not correspond to the sample solution or contradictions within the generated textual feedback. 
GPT-4 generated inconsistencies in about 22\,\% of the cases with a peak of 36\,\% in the third iteration of the \textit{Queue}. A notable example for the \textit{SimpleWhileLoop} is GPT-4 highlighting the initialization of \textit{i} at 0 as a ``Logical error'' saying \say{that it should begin at 1}. At the same time, it acknowledged that starting at 0 (as in the submitted code) ``would still produce the same correct sequence''. Similarly, for the \textit{Queue}, the feedback presents an inconsistency: GPT-4 recommends not to use getters and setters for the inner class, but recommends marking ``fields as private and accessing them through getters and setters''.

%redundancies
Redundancies (RD) were identified in about 10\,\% of the feedback outputs. For instance, some corrections or suggestions had already been part of the student's submission. Feedback was also deemed redundant when it involved repeated or trivial suggestions, e.\,g., \say{`QueueEntry' should be named `QueueEntry'}.

\section{Discussion}
The characterization of the feedback generated by GPT-4 shows that there are significant differences, compared to its previous version (GPT-3.5) and other LLMs. For example, the generated feedback is much longer, more structured, and got more complex. The median length is four times larger than reported in related work~\cite{LMU-TEL/ADS2023} (which used the same dataset). Studies on other models had reported issues with output formatting, such as an incorrect capitalization (cf. \cite{LMU-TEL/ADS2023,Finnie2022}). We cannot confirm these issues for GPT-4. 
Furthermore, applying the suggestions or using the provided model solution of the feedback always leads to a completely correct solution (except for two cases, cf. \autoref{sec:fbcorr}).
Previous work had reported significant misleading information with GPT-3.5~\cite{kiesler2023exploring}. \citeauthor{LMU-TEL/ADS2023} showed an accuracy of the correctness classification of 73\,\%. Here, GPT-4 reaches 84\,\% on the same data set~\cite{LMU-TEL/ADS2023}. \citeauthor{LMU-TEL/ADS2023} found significant differences in the feedback quality in response to fully correct, syntactically incorrect, and functionally incorrect student submissions. This does not seem to be the case for GPT-4 Turbo. In this work, 52\,\% of the feedback was fully correct and complete, which only applied to 31\,\% of the outputs generated by GPT-3.5 in prior work~\cite{LMU-TEL/ADS2023}. Yet, 48\,\% of the generated feedback is not complete and fully correct w.r.t. all details. Overall, the generated feedback seems to be quite consistent across the three runs.

A benefit of using LLMs is that it generates 100\,\% personalized feedback without the need to develop test cases. This is a crucial advantage compared to traditional e-assessment systems that mostly provide simple informative feedback generated by test cases or compiler error messages~\cite{jeuring2022towards}. The feedback by GPT-4 Turbo is more elaborated and always contained explanations and code. In contrast, GPT-3.5 not always offered code and text~\cite{LMU-TEL/ADS2023,kiesler2023exploring,kiesler2023large}. Another novelty compared to GPT-3.5 is that GPT-4 provides the output of students' code.
Hints on possible memory leaks (not interfering with functional correctness), rejecting functional correct implementations not using a singly-linked list, variable naming, content of comments, performance optimizations, simplifications of code, and coding style show the potential of GPT-4 for using it as a tool for formative assessment. Before the broad availability of generative AI, detecting such issues required manual inspection, white-box testing, or professional tools.

%Quality of the feedback:
As mentioned, the overall feedback is very detailed, long, and not always well ordered. 48\,\% of the generated feedback is incomplete and/or not fully correct, containing incorrect classifications, redundancies, inconsistencies, or problematic explanations. These aspects can make it more difficult for students to understand the feedback, increasing the cognitive load ~\cite{Sweller1994}. Similarly, the wording is not always appropriate for novices. Some comments in the feedback mention, for example, generics, concurrency, or improvements on the provided interface, which are likely to overwhelm novices who do not yet know these concepts.

The generated feedback almost always contains a model solution but rarely code snippets with gaps and instructions. 
Even if there is no consent among experts about feedback strategies (cf. \cite{jeuring2022towards}), this approach does not seem to encourage students to improve their submissions step-by-step. 
Another aspect worth mentioning is the absence of motivational statements, which is in contrast to related work~\cite{kiesler2023exploring} and, above all, to human tutors \cite{LMU-TEL/SH22}. GPT-4 Turbo also ignored a student's question as part of a code comment. A human tutor would likely not have done that.

To conclude, using GPT-4 Turbo for automatically generating feedback does not seem to be advisable. The same applies to students using it without guidance or prior instruction. Nevertheless, it may be used to support teaching assistants, or advanced students who understand basic concepts and thus the provided feedback. In practice, (malicious) prompt injections must also be prevented.

Further research should focus on the pedagogical integration of the feedback, its consistency, how it can be tailored to the prior knowledge of the students, and how it can be linked to (the progress of) a specific course. The error classifications provided by GPT-4 may also help build a student model as part of an adaptive learning system.
At the same time, the inherent dependency on OpenAI should be noted, if LLMs like GPT are used in educational settings. Sending student submissions to a third party may raise privacy concerns. Hence, locally installed or offline LLMs are worth further consideration and research.

\section{Threats to Validity} 

A limiting factor of this work is that OpenAI's model is under active development. Therefore, we documented when the experiment was conducted, which model was used, and how the output was generated. 
However, it should be noted that GPT-4 Turbo is designed to predict subsequent tokens from previous ones. This is why its answers can vary upon regeneration, even when presented with identical prompts and inputs. For this reason, each submission was submitted to GPT-4 three times. 
The obtained results may have also been influenced by other factors, such as the programming language and the task specifications, which can vary across institutions. 
Finally, the limitations of the qualitative research paradigm and the content analysis technique are acknowledged. To ensure an intersubjective understanding and reliability, all three authors were involved in the classification.

\section{Conclusions and Outlook}

%conclusion
In the context of large introductory programming classes and educators' limited resources for providing individual feedback, Large Language Models may be helpful for learners to provide feedback, e.\,g., when they are stuck. Due to the well-known challenges of LLMs (e.\,g., falsifications, ``lying''), however, it is crucial to evaluate its feedback characteristics before applying it in a course context with students or incorporating it into a learning system. 

Therefore, the present work explored and characterized GPT-4 Turbo's output when prompted with an introductory programming task (\textit{SimpleWhileLoop} and \textit{Queue}) and respective student solutions, which were selected from a dataset gathered in an introductory programming course. The qualitative thematic analysis of the generated feedback texts revealed that all of the generated feedback is personalized. Moreover, the application of all corrections and suggestions in the feedback would have resulted in achieving the fully correct solution -- except for two cases. However, only 52\,\% of the provided feedback was actually complete and fully correct in all details. In addition, the feedback provided actionable information on how to optimize the code and recommended stylistic changes for the majority of submissions. 

We conclude that GPT-4 provides significantly improved feedback compared to older versions, as it performs better (e.\,g., accuracy). It correctly recognizes output formatting and provides more structured feedback. At the same time, there are still issues such as misleading feedback, incorrect/\allowbreak{}problematic explanations for corrections, redundancies, and inconsistencies within a generated feedback, but also across all outputs.

%future work: 
The present work offers several pathways for future work, such as the evaluation of the feedback from a pedagogical perspective, how well it addresses learner's informational needs, and how to integrate specific feedback categories into learning environments or formative assessment systems. 

%\begin{acks}
We thank all the students for their consent to use their submissions for (this) research. 
This research is part of the project AIM@LMU funded by the German Federal Ministry of Education and Research (BMBF) under the grant number 16DHBKI013.
%\end{acks}

\balance

\bibliographystyle{ACM-Reference-Format}
\bibliography{main}

\end{document}